\documentclass{article}

\usepackage{amsmath,amsfonts,amssymb}
\usepackage{times}

\usepackage{graphicx}
\usepackage{natbib}
\usepackage{algorithm,algorithmic}
\usepackage{booktabs,multirow}
\usepackage{subfigure}
\usepackage{csquotes}

\usepackage{hyperref}

\usepackage[accepted]{whi2018}

\icmltitlerunning{Contrastive Explanations with Local Foil Trees}

\begin{document}

\twocolumn[
\icmltitle{Contrastive Explanations with Local Foil Trees}


\icmlsetsymbol{equal}{*}

\begin{icmlauthorlist}
\icmlauthor{Jasper van der Waa}{equal,tno,tud}
\icmlauthor{Marcel Robeer}{equal,tno,uu}
\icmlauthor{Jurriaan van Diggelen}{tno}
\icmlauthor{Matthieu Brinkhuis}{uu}
\icmlauthor{Mark Neerincx}{tno,tud}
\end{icmlauthorlist}

\icmlaffiliation{tno}{Perceptual and Cognitive Systems, Dutch Research Organization for Applied Research (TNO), Soesterberg, The Netherlands}
\icmlaffiliation{tud}{Interactive Intelligence group, Technical University of Delft, Delft, The Netherlands}
\icmlaffiliation{uu}{Department of Information and Computing Sciences, Utrecht University, Utrecht, The Netherlands}

\icmlcorrespondingauthor{Jasper van der Waa}{jasper.vanderwaa@tno.nl}

\icmlkeywords{Explainable Artificial Intelligence, Interpretable Machine Learning, Contrastive Explanations, Decision Trees, Model Agnostic, Machine Learning, IJCAI, Human Interpretable Machine Learning, WHI, XAI, iML, Foil}

\vskip 0.3in
]


\printAffiliationsAndNotice{\icmlEqualContribution} 

\begin{abstract}
Recent advances in interpretable Machine Learning (iML) and eXplainable AI (XAI) construct explanations based on the importance of features in classification tasks. However, in a high-dimensional feature space this approach may become unfeasible without restraining the set of important features. We propose to utilize the human tendency to ask questions like ``Why this output (the fact) instead of that output (the foil)?'' to reduce the number of features to those that play a main role in the asked contrast. Our proposed method utilizes locally trained one-versus-all decision trees to identify the disjoint set of rules that causes the tree to classify data points as the foil and not as the fact. In this study we illustrate this approach on three benchmark classification tasks.
\end{abstract}

\section{Introduction}
The research field of making Machine Learning (ML) models more interpretable is receiving much attention. One of the main reasons for this is the advance in such ML models and their applications to high-risk domains. Interpretability in ML can be applied for the following purposes: (i) transparency in the model to facilitate understanding by users \citep{Herman2017}; (ii) the detection of biased views in a model \citep{crawford2016, caliskan2017};  (iii) the identification of situations in which the model works adequately and safely \citep{barocas2016, coglianese2016, friedler2018}; (iv) the construction of accurate explanations that explain the underlying causal phenomena \citep{Lipton2016}; and (v) to build tools that allow model engineers to build better models and debug existing models \citep{kulesza2011, kulesza2015}.

The existing methods in iML focus on different approaches of how the information for an explanation can be obtained and how the explanation itself can be constructed. See for example for an overview the review papers of \citet{Guidotti2018} and \citet{Chakraborty2017}. A number of examples of common methods are: ordering the feature's contribution to an output \citep{Datta2016,Lei2016,Ribeiro2016}, attention maps and saliency of the features \citep{Selvaraju2016,Montavon2017a,Sundararajan2017,Zhang2017}, prototype selection, construction and presentation \citep{Nguyen2016}, word annotations \citep{Hendricks2016, Ehsan2017}, and summaries with decision trees \citep{Krishnan1999,Thiagarajan2016,Zhou2016} and decision rules \citep{Hein2017,Malioutov2017,Puri2017,Wang2017}. In this study we focus on feature-based explanations. Such explanations tend to be long when based on all features or use an arbitrary cutoff point. We propose a model-agnostic method to limit the explanation length with the help of contrastive explanations. The method also adds information of how that feature contributes to the output in the form of decision rules.

Throughout this paper, the main reason for explanations is to offer transparency in the model's given output based on which features play a role and what that role is. A few methods that offer similar explanations are LIME \citep{Ribeiro2016}, QII \citep{Datta2016}, STREAK \citep{Elenberg2017} and SHAP \citep{Lundberg2016}. Each of these approaches answers the question ``Why this output?'' in some way by providing a subset of features or an ordered list of all features, either visualized or structured in a text template. However, when humans answer such questions to each other they tend to limit their explanations to a few vital points \citep{Pacer2017}. This human tendency for simplicity also shows in iML: when multiple explanations hold we should pick the simplest explanation that is consistent with the data \citep{Huysmans2011}. The mentioned approaches do this by either thresholding the contribution parameter to a fixed value, presenting the entire ordered list or by applying it only to low-dimensional data. 

This study offers a more human-like way of limiting the list of contributing features by setting a contrast between two outputs. The proposed contrastive explanations present only the information that causes some data point to be classified as some class instead of another \citep{Miller2017a}. Recently, \citet{Dhurandhar2018} have proposed constructing explanations by finding contrastive perturbations---minimal changes required to change the current classification to any arbitrary other class. Instead, our approach creates \textit{contrastive targeted explanations} by first defining the output of interest. In other words, our contrastive explanations answer the question ``Why this output instead of that output?''. The contrast is made between the \textit{fact}, the given output, and the \textit{foil}, the output of interest. 

\begin{figure}[ht]
\vskip 0.2in
\begin{center}
\centerline{\includegraphics[width=0.9\columnwidth]{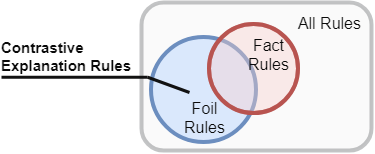}}
\caption{This figure shows the general idea of our approach to contrastive explanations. Given a set of rules that define data points as either the fact or foil, we take the relative complement of the fact rules in the foil rules to obtain a description how the foil differs from the fact in terms of features.}
\label{fig:rule_set_complement}
\end{center}
\vskip -0.2in
\end{figure}

A relative straightforward way to construct contrastive explanations given a foil based on feature contributions, is to compare the two ordered feature lists and see how much some feature differs in their ranking. However, a feature may have the same rank in both ordered lists but can be used in entirely different ways for the fact and foil classes. To mitigate this problem we propose a more meaningful comparison based on how a feature is used to distinct the foil from the fact. We train an arbitrary model to distinguish between fact and foil that is more accessible. From that model we distill two sets of rules; one used to identify data points as a fact and the other to identify data points as a foil. Given these two sets, we subtract the factual rule set from the foil rule set. This relative complement of the fact rules in the foil rules is used to construct our contrastive explanation. See Figure \ref{fig:rule_set_complement} for an illustration.

The method we propose in this study obtains this complement by training a one-versus-all decision tree to recognize the foil class. We refer to this decision tree as the Foil Tree. Next, we identify the fact-leaf---the leaf in which the current questioned data point resides. Followed by identifying the foil-leaf, which is obtained by searching the tree with some strategy. Currently our strategy is simply to choose the closest leaf to the fact-leaf that classifies data points as the foil class. The complement is then the set of decision nodes (representing rules) that are a parent of the foil-leaf but not of the fact-leaf. Rules that overlap are merged to obtain a minimum coverage rule set. The rules are then used to construct our explanation. The method is discussed in more detail in section \ref{sec:foil_trees}. An example of its usage is discussed in section \ref{sec:validation} on three benchmark classification tasks. The validation on these three tasks shows that the proposed method constructs shorter explanations than the fully feature list, provide more information of how these features contribute and that this contribution matches the underlying model closely. 

\section{Foil Trees; a way for obtaining contrastive explanations}
\label{sec:foil_trees}

The method we propose learns a decision tree centred around any questioned data point. The decision tree is trained to locally distinguish the foil-class from any other class, including the fact class. Its training occurs on data points that can either be generated or sampled from an existing data set, each labeled with predictions from the model it aims to explain. As such, our method is model-agnostic. Similar to LIME \citep{Ribeiro2016}, the sample weights of each generated or sampled data point depend on its similarity to the data point in question. Samples in the vicinity of the questioned data point receive higher weights in training the tree, ensuring its local faithfulness.

Given this tree, the `foil-tree', we search for the leaf in which the data point in question resides, the so called `fact-leaf'. This gives us the set of rules that defines that data point as the not-foil class according to the foil-tree. These rules respect the decision boundary of the underlying ML model as it is trained to mirror the foil class outputs. Next, we use an arbitrary strategy to locate the `foil-leaf'---for example the leaf that classifies data point as the foil class with the lowest number of nodes between itself and the fact-leaf. This results in two rule sets, whose relative complement define how the data point in question differs from the foil data points as classified by the foil-leaf. This explanation of the difference is done in terms of the input features themselves.

\begin{figure*}[ht]
\vskip 0.05in
\includegraphics[width=\textwidth]{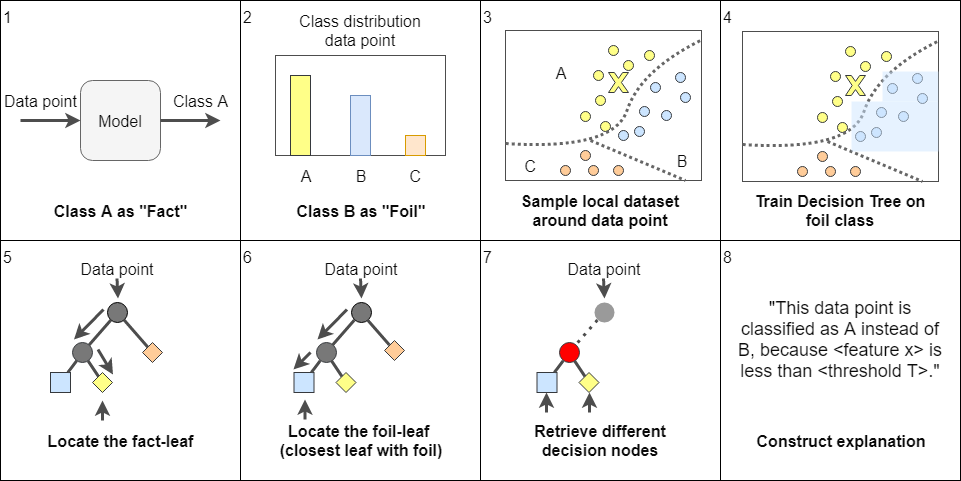}
\vskip -0.07in
\caption{The steps needed to define and train a Foil Tree and to use it to construct a contrastive explanation. Each step corresponds with the listed steps in section \ref{sec:foil_trees}.}
\label{fig:foil_tree_method}
\vskip -0.15in
\end{figure*}

In summary, the proposed method goes through the following steps to obtain a contrastive explanation for an arbitrary ML model, the questioned data point and its output according to that ML model:
\begin{enumerate}
\item \textbf{Retrieve the fact}; the output class.
\item \textbf{Identify the foil}; explicitly given in the question or derived (e.g. second most likely class).
\item \textbf{Generate or sample a local data set}; either randomly sampled from an existing data set, generated according to a normal distribution, generated based on marginal distributions of feature values or more complex methods.
\item \textbf{Train a decision tree}; with sample weights depending on the training point's proximity or similarity to the data point in question.
\item \textbf{Locate the `fact-leaf'}; the leaf in which the data point in question resides.
\item \textbf{Locate a `foil-leaf'}; we select the leaf that classifies data points as part of the foil class with the lowest number of decision nodes between it and the fact-leaf. 
\item \textbf{Compute differences}; to obtain the two set of rules that define the difference between fact- and foil-leaf, all common parent decision nodes are removed from each rule sets. From the decision nodes that remain, those that regard the same feature are combined to form a single literal.
\item \textbf{Construct explanation}; the actual presentation of the differences between the fact-leaf and foil-leaf.
\end{enumerate}
Figure \ref{fig:foil_tree_method} illustrates the aforementioned steps. The search for the appropriate foil-leaf in step 6 can vary. In Section \ref{sec:strategies} we discuss this more in detail. Finally, note that the method is not symmetrical. There will be a different answer on the question ``Why class A and not B?'' then on ``Why class B and not A?'' as the foil-tree is trained in the first case to identify class B and in the second case to identify class A. This is because we treat the foil as the expected class or the class of interest to which we compare everything else. In addition, even if the trees are similar, the relative complements of their rule sets are reversed

\subsection{Foil-leaf strategies}
\label{sec:strategies}
Up to now we mentioned one strategy to find a foil-leaf, however multiple strategies are possible---although not all strategies may result in a satisfactory explanation according to the user. The strategy used in this study is simply the first leaf that is closest to the fact-leaf in terms of number decision nodes, resulting in a minimal length explanation.

A disadvantage of this strategy is its ignorance towards the value of the foil-leaf compared to the rest of the tree. The nearest foil-leaf may be a leaf that classifies only a relatively few data points or classifies them with a relatively high error rate. To mitigate such issues the foil-leaf selection mechanism can be generalized to a graph-search from a specific (fact) vertex to a different (foil) vertex while minimizing edge weights. The foil-tree is treated as a graph whose decision node and leaf properties influence some weight function. This generalization allows for a number of strategies, and each may result in a different foil-leaf. The strategy used in this preliminary study simply reduces to each edge having a weight of one, resulting in the nearest foil-leaf when minimizing the total weights.

As an example, an improved strategy may be where the edge weights are based on the relative accuracy of a node (based on its leaves) or leaf. Where a higher accuracy results in a lower weight, allowing the strategy to find more distant, but more accurate, foil-leaves. This may result in relatively more complex and longer explanations, which nonetheless hold in more general cases. For example the nearest foil-leaf may only classify a few data points accurately, whereas a slightly more distant leaf classifies significantly more data points accurately. Given the fact that an explanation should be both accurate and fairly general, this proposed strategy may be more beneficial \citep{Craven1999}.

Note that the proposed method assumes the knowledge of the used foil. In all cases we take the second most likely class as our foil. Although this may be an interesting foil it may not be the contrast the user actually wants to make. Either the user makes its foil explicit or we introduce a feedback loop in the interaction that allows our approach to learn which foil is asked for in which situations. We leave this for future work.

\section{Validation}
\label{sec:validation}
The proposed method is validated on three benchmark classification tasks from the UCI Machine Learning Repository \citep{UCI}; the Iris data set, the PIMA Indians Diabetes data set and the Cleveland Heart Disease data set. The first data set is a well-known classification task of plants based on four flower leaf characteristics with a size of $150$ data points and three classes. The second data set is a binary classification task whose task is to correctly diagnose diabetes and contains $769$ data points and has nine features. The third data set is aims at classifying the risk of heart disease from no presence ($0$) to presence ($1\textrm{--}4$), consisting of $297$ instances with $13$ features.

To show the model-agnostic nature of our proposed method we applied four distinct classification models to each data set: a random forest, logistic regression, support vector machine (SVM) and a neural network. Table \ref{table:validation_results} shows for each data set and classifier the $F_1$ score of the trained model. We validated our approach on four measures; explanation length, accuracy, fidelity and time. These measures for evaluating iML decision rules are adapted from \citet{Craven1999}, where the mean length serves as a proxy measure demonstrating the relative explanation comprehensibility \citep{Doshi-Velez2017}. The fidelity allows us to state how well the tree explains the underlying model, and the accuracy tells us how well its explanations generalize to unseen data points. Below we describe each in detail:
\begin{enumerate}
\item \textbf{Mean length}; average length of the explanation in terms of decision nodes. The ideal value is in the range $[1.0,$ Nr. features$)$, since a length of $0$ means that no explanation is found and a length near the number of features offers little gain compared to showing the entire ordered feature contribution list as in other iML methods. 
\item \textbf{Accuracy}; $F_1$ score of the foil-tree for its binary classification task on the test set compared to the true labels. This measure indicates how general the explanations generated from the Foil Tree are on an unseen test set.
\item \textbf{Fidelity}; $F_1$ score of the foil-tree on the test set compared to the model output. This measure provides a quantitative value of how well the Foil Tree agrees with the underlying classification model it tries to explain.
\item \textbf{Time}; number of seconds needed on average to explain a test data point.
\end{enumerate}

Each measure is cross-validated three times to account for randomness in foil-tree construction. These results are shown in their respective columns in Table \ref{table:validation_results}. They show that on average the Foil Tree is able to provide concise explanations, with a mean length $1.33$, while accurately mimicking the decision boundaries used by the model with a mean fidelity of $0.93$ and generalizes well to unseen data with a mean accuracy of $0.92$. The foil-tree performs similar to the underlying ML model in terms of accuracy. Note that for the random forest, logistic regression and SVM models on the diabetes data set rules of length zero were found---i.e. no explanatory differences were found between facts and foils in a number of cases---, resulting in a mean length of less than one. For all other models our method was able to find a difference for every questioned data point.

To further illustrate the proposed method, below we present a single explanation of two classes of the Iris data set in a dialogue setting;

\begin{displayquote}
\begin{itemize}
    \item System: The flowertype is `Setosa'.
    \item User: Why `Setosa' and not `Versicolor'?
    \item System: Because for it to be `Versicolor' the `petal width (cm)' should be smaller  and the `sepal width (cm)' should be larger.
    \item User: How much smaller and larger?
    \item System: The `petal width (cm)' should be smaller than or equal to $0.8$ and the `sepal width (cm)' should be larger than $3.3$.
\end{itemize}
\end{displayquote}

The fact is the `Setosa' class, the foil is the `Versicolor' class and the total length of the explanation contains two decision nodes or literals. The generation of this small dialogue is based on text templates and fixed interactions for the user.


\begin{table*}[ht]
\caption{Performance of foil-tree explanations on the Iris, PIMA Indians Diabetes and Heart Disease classification tasks. The column 'Mean length' also contains the total number of features for that data set as the upper bound of the explanation length.}
\label{table:validation_results}
\vskip 0.15in
\begin{center}
\begin{small}
\begin{sc}
\begin{tabular}{llccccc}
 \toprule
Data set & Model & $F_1$ Score & Mean length &  Accuracy & Fidelity & Time\\
\midrule
\multirow{4}{*}{Iris}		& Random Forest 	    & 0.93 & 1.94 (4) & 0.96 & 0.97 & 0.014 \\
							& Logistic Regression 	& 0.93 & 1.50 (4) & 0.89 & 0.96 & 0.007 \\
							& SVM				    & 0.93 & 1.37 (4) & 0.89 & 0.92 & 0.010 \\
							& Neural Network		& 0.97 & 1.32 (4) & 0.87 & 0.87 & 0.005 \\
\midrule
\multirow{4}{*}{Diabetes} 	& Random Forest		    & 1.00 & 0.98 (9) & 0.94 & 0.94 & 0.041 \\
							& Logistic Regression	& 1.00 & 0.98 (9) & 0.94 & 0.94 & 0.032 \\
							& SVM				    & 1.00 & 0.98 (9) & 0.94 & 0.94 & 0.034 \\
							& Neural Network		& 1.00 & 1.66 (9) & 0.99 & 0.99 & 0.009 \\
\midrule
\multirow{4}{*}{Heart Disease} & Random Forest		& 0.94 & 1.32 (13) & 0.88 & 0.90 & 0.106 \\
							& Logistic Regression	& 1.00 & 1.21 (13) & 0.99 & 0.99 & 0.006 \\
							& SVM				    & 1.00 & 1.19 (13) & 0.86 & 0.86 & 0.012 \\
							& Neural Network		& 1.00 & 1.56 (13) & 0.92 & 0.92 & 0.009 \\
\bottomrule
\end{tabular}
\end{sc}
\end{small}
\end{center}
\vskip -0.15in
\end{table*}

\section{Conclusion}
Current developments in Interpretable Machine Learning (iML) created new methods to answer ``Why output A?'' for Machine Learning (ML) models. A large set of such methods use the contributions of each feature used to classify A and then provides either a subset of feature whose contribution is above a threshold, the entire ordered feature list or simply apply it only to low-dimensional data.

This study proposes a novel method to reduce the number of contributing features for a class by answering a contrasting question of the form ``Why output A (fact) instead of output B (foil)?'' for an arbitrary data point. This allows us to construct an explanation in which only those features play a role that distinguish A from B. Our approach finds the contrastive explanation by taking the complement set of decision rules that cause the classification of A in the rule set of B. In this study we implemented this idea by training a decision tree to distinguish between B and not-B (one-versus-all approach). A fact-leaf is found in which the data point in question resides. Also, a foil-leaf is selected according to a strategy where all data points are classified as the foil (output B). We then form the contrasting rules by extracting the decision nodes in the sub-tree from the lowest common ancestor between the fact-leaf and foil-leaf, that hold for the foil-leaf but not for the fact-leaf. Overlapping rules are merged and eventually used to construct an explanation.

We introduced a simple and naive strategy of finding an appropriate foil-leaf. We also provided an idea to extend this method with more complex and accurate strategies, which is part of our future work. We plan a user validation of our explanations with non-experts in Machine Learning to test the satisfaction of our explanations. In this study we tested if the proposed method is viable on three different benchmark tasks as well as to test its fidelity on different underlying ML models to show its model-agnostic capacity.

The results showed that for different classifiers our method is able to offer concise explanations that accurately describe the decision boundaries of the model it explains.

As mentioned, our future work will consist out of extending this preliminary method with more foil-leaf search strategies as well as applying the method to more complex tasks and validating its explanations with users. Furthermore, we plan to extend the method with an adaptive foil-leaf search to adapt explanations towards a specific user based on user feedback.


\bibliography{main}
\bibliographystyle{icml2018}

\end{document}